\def\year{2019}\relax
\documentclass[letterpaper]{article}
\usepackage{aaai19}  
\usepackage{times}  
\usepackage{helvet}  
\usepackage{courier}  
\usepackage{url}  
\usepackage{graphicx}  
\usepackage[draft,inline,nomargin,index]{fixme}
\usepackage[utf8]{inputenc}
\usepackage[T1]{fontenc}
\usepackage{color,amsfonts}
\usepackage{hyperref,booktabs}
\usepackage{latexsym,amsmath,nicefrac,amssymb}
\usepackage{microtype}
\usepackage{framed}
\usepackage{multirow}
\usepackage[linesnumbered,ruled]{algorithm2e}
\usepackage{subcaption}
\usepackage{wrapfig}
\usepackage{enumitem}

\FXRegisterAuthor{tl}{Taesung}{\color{blue}Taesung}
\FXRegisterAuthor{im}{Ian}{\color{magenta}Ian}
\FXRegisterAuthor{be}{Ben}{\color{red}Ben}
\FXRegisterAuthor{ds}{Dong}{\color{cyan}Dong}

\newcommand{\eg}{{\it e.g.}}%
\newcommand{\ie}{{\it i.e.}}%
\newcommand{\CYJ}[1]{}

\def\sec{Section}
\def\tab{Table}

\def\fig{Figure}

\newcommand{\sample}{{\ensuremath{\mathsf{Sample}}}\xspace}
\newcommand{\argmaxsample}{{\ensuremath{\mathsf{Argmax Sample}}}\xspace}
\newcommand{\random}{{\ensuremath{\mathsf{Random}}}\xspace}
\newcommand{\jacobian}{{\ensuremath{\mathsf{Jacobian\ Augmentation}}}\xspace}

\title{Defending Against Machine Learning Model Stealing Attacks Using Deceptive Perturbations}

\author{
Taesung Lee${}^1$\qquad Benjamin Edwards${}^2$ \qquad Ian Molloy${}^3$ \qquad Dong Su${}^4$\\
IBM Research AI\\
\texttt{\{${}^1$taesung.lee, ${}^2$benjamin.edwards, ${}^4$dong.su\}@ibm.com} \\
\texttt{${}^3$molloyim@us.ibm.com} 
}

\begin{document}
\maketitle

\begin{abstract}
Machine learning architectures are readily available on the web,
but creating the high quality training data is costly.
However, a pretrained model on a cloud service 
can be used to generate labeled data to 
steal the model if the adversary can obtain output labels for chosen inputs.
To protect against these attacks,
it has been proposed to limit the information provided to the adversary
by omitting probability scores,
significantly impacting the utility of the provided service.
In this work, we illustrate how a service provider can still provide
useful, albeit misleading, class probability information,
while significantly limiting the success of the attack.
Our defense forces the adversary to discard
the class probabilities, requiring significantly
more queries before they can train a model with comparable performance.
We evaluate several attack strategies, model architectures,
and hyperparameters under varying adversarial models,
and evaluate the efficacy of our defense against the strongest adversary.
Finally, we quantify the amount of noise injected into
the class probabilities to measure the loss in utility,
\eg, adding 1.26 nats per query on CIFAR-10 and 3.27 on MNIST.
Our evaluation shows
our defense can degrade the accuracy of the stolen model at least 20\%,
or require up to 64 times more queries
while keeping the accuracy of the protected model almost intact.

\end{abstract}

\section{Introduction}
\label{sec:intro}

The success of neural networks has resulted in many web services based on them,
including services providing APIs to label input samples for small sums of money.
Many state-of-the-art neural network models are readily available
in the literature or online,
and unlabeled data (\eg, images and corpus) are also often abundant on the web.
But labeling data to train a machine learning model
is expensive, difficult and error-prone even for simple tasks~\cite{ringger2008assessing}.
It is even more difficult for domains requiring expert knowledge
(\eg, coreference resolution in the medical domain).
However, once an adversary acquires enough labels using the web service,
the attacker can replicate the neural network and no longer needs
to pay for the service~\cite{florian2016}.
For example, current image classification services charge around \$1--\$10 per 1,000 queries, depending on the sophistication or customization of the model~\cite{google_price,azure_price,watson_vr_price}.
Moreover, replicating neural networks greatly expands the attack surface,
allowing white-box attacks against black-box services~\cite{papernot2017jacobian}.


In particular, we consider a scenario that an attacker uses
the probability values returned by the base model on the cloud service
to boost the model stealing process.
The cloud service often provide probability values to show confidence.
When stealing models, \cite{florian2016} claim that using probabilities
instead of labels alone reduces the number of required samples by 50--100$\times$.
We also confirm that using probabilities
can improve the convergence, and also increases the converged model accuracy
in \sec~\ref{sec:exp:attack} and \ref{sec:exp:variants}.

To mitigate this problem, we propose to add smart noise in the output probability
that maintains the output class label of the model not to harm the accuracy.
We aim to force the attacker to discard the probability and use labels only,
which is a lower bound of the optimal attack for an accuracy-preserving defense.
To evaluate the performance, we consider two types of attacks.
First, we identify and test an attack that can replicate an unprotected model quickly.\footnote{We observe that a defense-aware attack performs worse on an unprotected model.}
Our evaluation considers diverse datasets, attack parameters, neural networks architectures
and domains (images and text),
and shows that our approach can degrade the stolen model accuracy by 20\% or more
while keeping the protected model accuracy almost intact.
Second, we consider defense-aware attacks,
including exploiting the same defense layer,
reversing the noise, using a different loss function, and using only labels.
With the accuracy-preserving defense,
the attacker can still get correct labels by applying argmax to the probability vector,
and therefore using only labels is a lower bound of the best attack,
which is much slower to converge.
We aim to force the attacker to use this suboptimal attack by eliminating better attacks
and show diverse defense-aware attacks fail to achive better accuracy than the lowerbound.

\section{Related work} 
\label{sec:related}

The problem of inferring secret model parameters
by observing the output classification to a given set of inputs
has been recently studied.
\cite{florian2016} claim that
using high-precision confidence values and class labels
obtained from a machine learning cloud service,
their attack can steal the base model of
several types including decision trees, logistic regression,
support vector machines and simple neural networks.
For simple parametric models such as logistic regression,
they solve a linear system from the obtained probabilities.
For decision trees, they develop a path-finding algorithm
to exploit the confidence value as pseudo-identifiers
for paths in the tree to discover the tree structure.
For neural networks, they leverage a method we denote by \sample{}
that uses a set of samples and query a randomly drawn batch,
and train the network with the output from the base model.
\cite{stealingReg} also consider stealing
a machine learning parameter, but their work is limited to
the regularization parameter, not the entire model.

\cite{papernot2017jacobian} proposed
stealing a black-box neural network model to generate adversarial examples.
They assume the attacker has a limited number of training data,
and propose to use 
a Jacobian-based heuristic in order to find examples defining
the decision boundary of the target model.
We extend the analyses of \sample{} and \jacobian{} 
with five datasets, four neural network architectures
as well as other attack methods to leverage in our defense evaluation.
To our knowledge, this is the first study
mitigating such model stealing attacks.

Student-teacher models have been used to compress
a sophisticated teacher machine learning model into a smaller student model with less parameters~\cite{bucila2006model,Romero2014FitNetsHF}.
Using the output probability vectors from the teacher to train the student,
we can obtain a similarly performing student model with far less parameters.
This paradigm focuses only on improving the student model 
with less parameters, unlike our goal of preventing it,
and tries to leverage more information about the teacher (white-box)
to better train or design the student,
which is inapplicable in our cloud service scenario.

\section{Methods} \label{sec:method}


In this section, we propose an add-on layer that can be applied to
most neural network classifiers to protect against model stealing from cloud service APIs.
This layer adds a small controllable perturbation
\emph{maximizing the loss} of the stolen model while preserving the accuracy.
That is, instead of attempting to detect an attack,
we apply noise that has little influence to normal users,
but still degrades and slow down the model stealing attack.
An optimal defense should provide utility to the service consumers, while providing no measurable benefit to the adversary beyond a final label.

We consider neural networks that extract features from the input data
and aggregate them throughout the layers to generate class probabilities of the input.
Typically, the last layer is an activation function
producing probability values ranging from 0 to 1 and sum to 1,
given \emph{logits}, the unbounded vector from the previous layer.
In most cases, the softmax function without parameters is used.
That is, most neural network classifiers of $K$ classes can be represented as
$y = f(x) = \sigma(g(x))$, mapping input $x$ to output $y$,
where $g(\cdot)$ is a function to a $K$-dimensional real vector,
and $\sigma(\cdot)$ is a normalization function (\eg, softmax)
mapping a vector to $K$ probability values summing to 1.

An attacker with samples $X=\{x^i\}$ can query a neural network
on the remote server (\emph{base model}) to obtain
the corresponding pseudo-labels $Y=\{y^i\}$,
and train their own neural network (\emph{stolen model}).
The completely replicated network should have the minimum loss $L$ with respect to $Y$,
which is usually defined using the cross entropy loss function.
That is,
\begin{equation}
L(X,Y) = - \sum_i \sum_j y^i_j \log f(x^i)_j
\end{equation}
where $y^i_j$ and $f(x^i)_j$ represent $j$-th dimension of $y^i$ and $f(x^i)$, respectively.

In this setting, we propose to add noise to the server response $\{y^i\}$
that results in a high loss so that the attacker's optimizer ill-trains the network.
Toward this goal, we first assume that the stolen model
already perfectly replicated the base model,
and parameterize the possible perturbation of $y^i$ with conditions
to maximize loss $L$.

In particular, the perturbed probability vector $\hat{y}^i$ should have the following properties.
First, the sum across the dimensions must be 1.
Second, we should be able to control the magnitude of the perturbation.
Last, the accuracy should be preserved, \ie,
$\hat{y}^i_k \geq \hat{y}^i_j$ for $j=1,\ldots,K$ and $k$ such that $y^i_k \geq y^i_j$.
To this goal, we consider additive perturbation with normalization:
$\hat{y}^i_j = \alpha^i (y^i_j - r(y^i_j))$
where $\alpha^i$ is a sum-to-1 normalizer for $y^i$,
and $r(y^i_j)$ is the noise function we seek with the following parameterization:
\begin{equation}
r(y^i_j) = \beta(s(z^i_j) - 1/2)
\label{eq:noise_func}
\end{equation}
where
$s(\cdot)$ is a sigmoid function,
$z^i_j$ is a perturbation,
and $\beta$ is a positive magnitude parameter;
with a constraint
$\hat{y}^i_k \geq \hat{y}^i_j$ for $j=1,\ldots,K$ preserving the accuracy.
Using a derivative test, we can find $L(x, \hat{y})$ has critical points
when $z^i_j \rightarrow \pm \inf$.
In particular, $L(x,\hat{y})$ is maximized
when $z^i_k \rightarrow \inf$ for $k$ such that $y^i_k \geq y^i_j$ for $j=1,...\ldots,K$,
and $z^i_j \rightarrow -\inf$ for $j$ with low $y^i_j$.

\begin{figure}[htbp]
	\centering
		\includegraphics[width=1\linewidth]{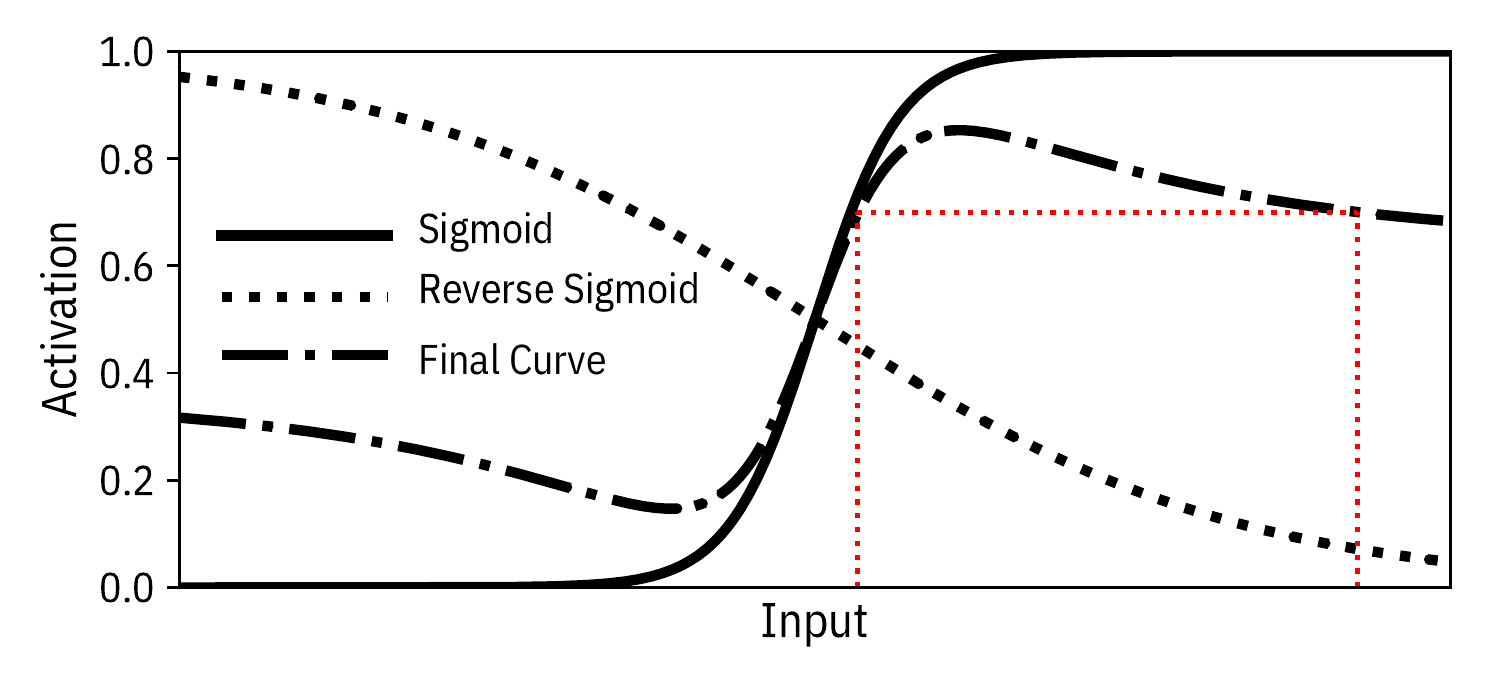}
	\caption{Example Reverse Sigmoid and Sigmoid activation functions.}
	\label{fig:rsig}
\end{figure}

Instead of directly setting $z^i_j$ to maximize the loss,
we use a heuristic approximation to $+\inf$ when $y^i_k$ is the largest
and $-\inf$ otherwise, not to completely lose the probability values:
$z^i_j = \gamma s^{-1}(y^i_j)$
where $\gamma$ is a positive dataset and model specific convergence parameter,
and $s^{-1}(y^i_j)$ is the pseudo-logit of $y^i_j$ that amplifies the behavior of $y^i_j$
and makes the perturbation comparable to the original probability.
With this approximation, we obtain \emph{Reverse Sigmoid} perturbation $r(y^i_j)$:
\begin{equation}
r(y^i_j) \approx \beta(s(\gamma s^{-1}(y_j^i)) - 1/2)
\end{equation}
This function has a shape of flipped sigmoid function as shown in \fig~\ref{fig:rsig},
and the final perturbed probability value is computed as follows:
\begin{equation}
\hat{y}^i_j = \alpha^i \left( y^i_j - \beta(s(\gamma s^{-1}(y_j^i)) - 1/2) \right)
\end{equation}
This function has humps that prevent simple inversion impossible as depicted in \fig~\ref{fig:rsig}.

The main advantage of using Reverse Sigmoid $s(\gamma s^{-1}(y^i_j))$ in $r(y^i_j)$ is two-fold.
First, we don't completely lose the meaning of the probability values,
in contrast to always returning the same values for top-1 and bottom-1 classes.
Second, this function form adds ambiguity that prevents a simple inversion.
As we can see in \fig~\ref{fig:rsig}, the final deceptive probability curve
has two $y^i_j$ values that have the same $\hat{y}^i_j$ in the range
$[0,\frac{\alpha^i\beta}{2}]$ and $[\alpha^i(1-\frac{\beta}{2}), 1]$,
except where the first derivative is zero,
making the exact inverse function impossible
and the inversion difficult. 
The adversary cannot attempt to select from each value because it results in an exponential number of options, up to $2^K$ possibilities \emph{per} sample.
For the same reason,
the network cannot be replicated perfectly
even if the attacker exploit Reverse Sigmoid
in the attacker's model
or leverage mean squared errors instead of cross loss entropy,
which we show empirically in \sec~\ref{sec:exp:variants}.


\section{Experiments}

In this section, we evaluate the proposed defense method.
For this goal, we consider two types of attacks.
We first identify the best attack for an undefended model,
including query generation, parameters, and underlying models in \sec~\ref{sec:exp:attack}.
For an undefended model, we find that the best performing attack designed for a defended model
does not perform as good as that for an undefended model.
Then, we evaluate the defense methods on five image datasets and one text dataset
with five measures using the identified attack in \sec~\ref{sec:exp:defense},
and we discuss relations of the base models and the stolen models in \sec~\ref{sec:exp:add}.
We further evaluate the defense against possible attacks
when the attacker knows more information about the defense in \sec~\ref{sec:exp:variants},
and show these attacks cannot achieve better performance than using only labels,
which is a lower bound of the best attack for the accuracy-preserving defense.
Basic data augmentation (shift/flip) is applied in all training~\cite{dataAugmentation}.
MobileNet~\cite{MobileNet} and Xception~\cite{Xception} are optimized with Adam optimizer~\cite{adamOptimizer},
AllConv is optimized with the vanilla stochastic gradient descent as in its original paper~\cite{AllConvNet},
and RMSProp~\cite{RMSProp} is used otherwise to achieve the best performance for each individual model.

We use the following five measures to evaluate the performance of attacks and defenses.
\begin{itemize}
\item Agreement: Top-1 model accuracy of the stolen model treating the base model as ground truth.
\item Cosine: The average cosine similarity of output probability vectors of the stolen and the base models. 
\item MAE (Mean absolute error): The average absolute errors of the predictions of the stolen and the base models per class.
\item KL-divergence: KL-divergence between the probabilities of the stolen and the base models.
\item Accuracy: The prediction accuracy.
\end{itemize}
The first four measures do not use labels of the test dataset,
and focus on the intrinsic model replicability.
Accuracy focuses on the extrinsic performance on tasks.

We use six datasets including IMDB sentiment classification~\cite{IMDB}, MNIST~\cite{MNIST},
FASHION-MNIST~\cite{FASHIONMNIST}, CIFAR-10, and CIFAR-100~\cite{CIFAR}, and STL-10~\cite{STL10}.
These classification datasets cover different degrees of difficulty,
and a model for an easy dataset is easier to attack and harder to defend as shown in \sec~\ref{sec:exp:defense}.
If we can defend them, it's likely we can defend larger models and sophisticated datasets as well. 
For the image datasets, we randomly hold off 33\% of the original training data, uniformly from each class,
to assign a portion of them to the attacker after removing the labels and the probability values.
Note that this partitioning will
allow only 67\% of the original training data for the base model on the cloud,
and result in a small drop in model accuracy
for each of the base models compared to the state-of-the-art.
For the text dataset, we split the test data in the same way due to the small training data.

\subsection{Threat Model and Attacks} \label{sec:exp:attack}

For the evaluation, we consider the following adversarial model.
The adversary knows the architecture of the model being attacked,
has a given number of unlabeled input samples (\# samples),
and can send a fixed number of (adaptive) queries
to the base model to obtain labels and probability values.
We do not assume the adversary is computationally bounded for training. 

We compare the following attack strategies to generate queries
and use the response as the training data.
\begin{itemize}
\item \sample{}: The attacker has a certain number of data samples in hand,
and queries them to the base model.
\item \argmaxsample{}: Same as \sample{}, but this attack uses the top-1 class label only.
\item \random{}: The attacker does not rely on any existing data samples,
and instead generates uniform random queries.
The attacker knows the ranges and the dimensions of the input values.
\item \jacobian{}: The attacker has a certain number of data samples,
and generates more samples using the Jacobian method~\cite{papernot2017jacobian}.
The stolen model is trained using the response from the base model 
for both the data samples, and the generated samples.
The trained model is further used to generate more samples.
We set $\lambda=0.1$ and substitute training epochs $\rho=40$.
\end{itemize}

The attacks also use randomized image augmentation that shifts,
and/or flips images during the training\footnote{Flipping is not applied on the MNIST dataset.}.
As a neural network is trained for multiple epochs,
the same training sample is used multiple times.
Adding slight change to the image in every epoch provides much better generalization power,
and results in better test accuracy~\cite{dataAugmentation}.

We compare these attacks with various parameters,
identify the attack most successful at replicating the base model,
and evaluate our defenses against the strongest attack.
%
We use a simple convolutional network, denoted by Simple, for both the base and the stolen models:
64 $3 \times 3$ conv, 64 $3 \times 3$ conv, $2 \times 2$ max pooling,
128 $3 \times 3$ conv, 128 $3 \times 3$ conv, $2 \times 2$ max pooling,
256 dense, 256 dense, and softmax layers.


\begin{figure*}[t!]
\begin{minipage}{0.75\textwidth}
  \centering
  \includegraphics[width=1.0\linewidth]{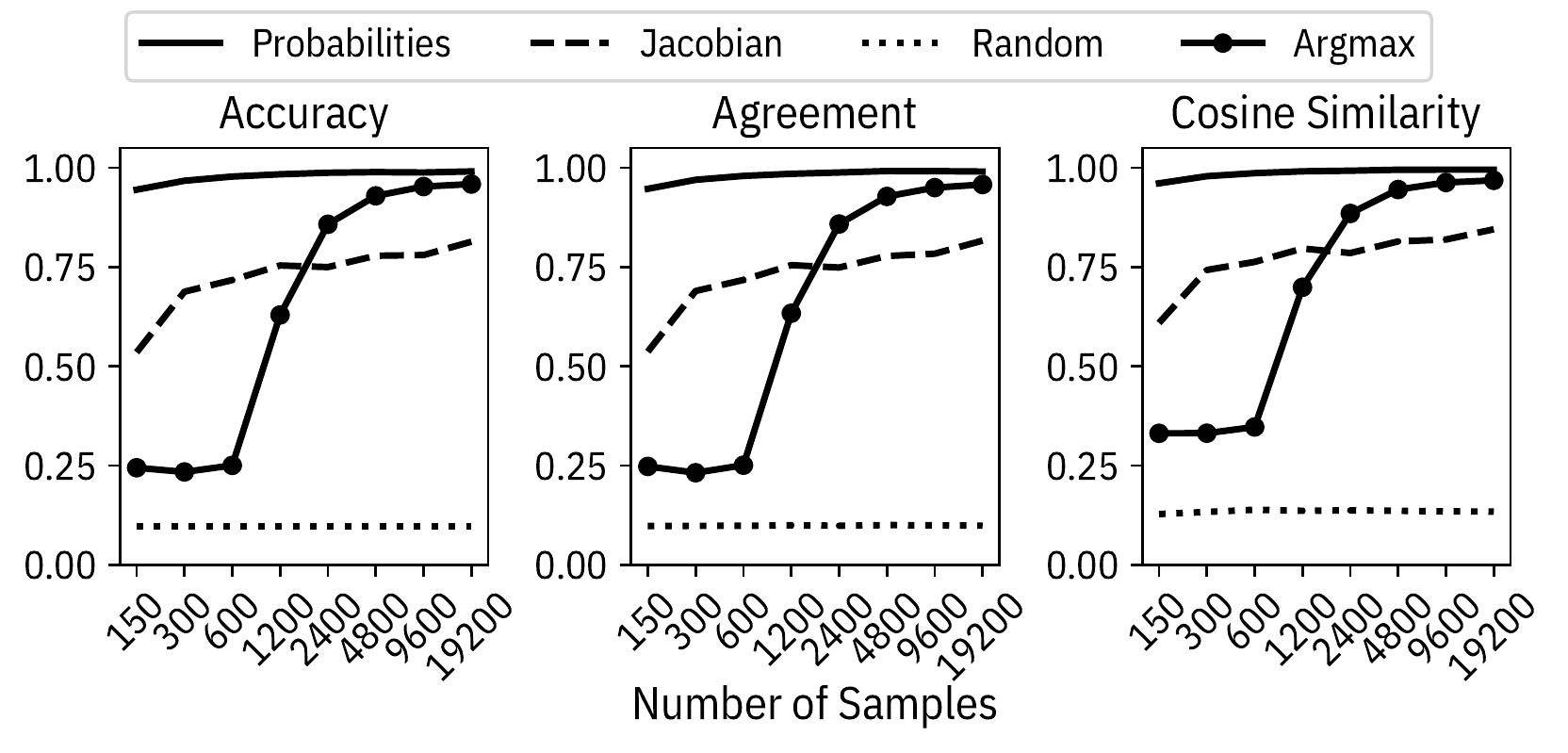}
  \caption{Performance of three different model stealing approaches.}
  \label{fig:different_attacks}
\end{minipage}
\begin{minipage}{0.2\textwidth}
	\centering
		\addtolength{\tabcolsep}{-5pt}
		\vspace{15pt}
		\small
		\begin{tabular}{ccc}
		\includegraphics{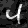} & \includegraphics{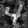} & \includegraphics{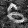} \\
		3 & 5 & 3\\
		\includegraphics{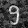} & \includegraphics{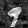} & \includegraphics{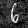} \\
		8 & 8 & 8 \\
		\includegraphics{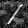} & \includegraphics{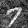} & \includegraphics{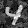} \\
		2 & 2 & 1
		\end{tabular}
		\vspace{15pt}
		\caption{Jacobian augmented inputs}
		\label{fig:attacks:jacobian}
\end{minipage}
\end{figure*}


\paragraph{Query Generation}
The most limiting resource for an adversary is real data samples to query the base model
(\eg, medical records, employee faces).
Therefore, we evaluate the attacker methods on varying \# samples first,
and set attacker budget to 50,000 queries, and training steps to 16,000 with 64-sized batch.
\fig~\ref{fig:different_attacks} shows the performance of the three attack strategies
against the number of samples on the MNIST dataset.
We can see that \sample{} attack performs best for various \# samples.
Like all other attack methods, \sample{} leverage data augmentation
to generate more samples.
Our preliminary experiment showed that using data augmentation
significantly improved the test accuracy (approx. 85\% increase with 150 samples).
\argmaxsample{} also steals the model with high accuracy,
but its replication is much slower especially at the beginning.
For example, with data augmentation, \sample{} reaches 97\% agreement with 300 samples,
but \argmaxsample{} goes only up to 96\% with 19200 samples
($\approx 64 \times$ samples).

The Jacobian augmentation is used to push samples towards the boundaries
of each class in the direction of greatest increase in the loss function
to generate samples defining the decision boundary.
When applied to the input image, it successfully probes
the classification boundary of the model, but results in
generating imperceptible perturbation to an image
so that the neural network misclassifies the input~\cite{papernot2016limitations}.
Thus, using Jacobian generates adversarial examples
that are misclassified by the base model
(\eg, \fig~\ref{fig:attacks:jacobian} are all misclassified as indicated).
Although they are useful in adversarial example generation scenario~\cite{papernot2017jacobian},
leveraging these labels from the base model results
in teaching the replicated model the wrong classification.

On the other hand, although \random{} attack has high training accuracy,
its performance on test data is poor, giving the accuracy of random guess.
Most of samples generated by \random{} falls in one class,
and this leads to train the replicated model to predict just one class
regardless the input.
This shows the importance having legitimate data samples to query the model.
Based on these results, we use \sample{} attack,
which can most accurately replicate the model in the defense-unaware scenario.


\begin{figure*}[t!]
  \centering
  \includegraphics[width=0.9\linewidth]{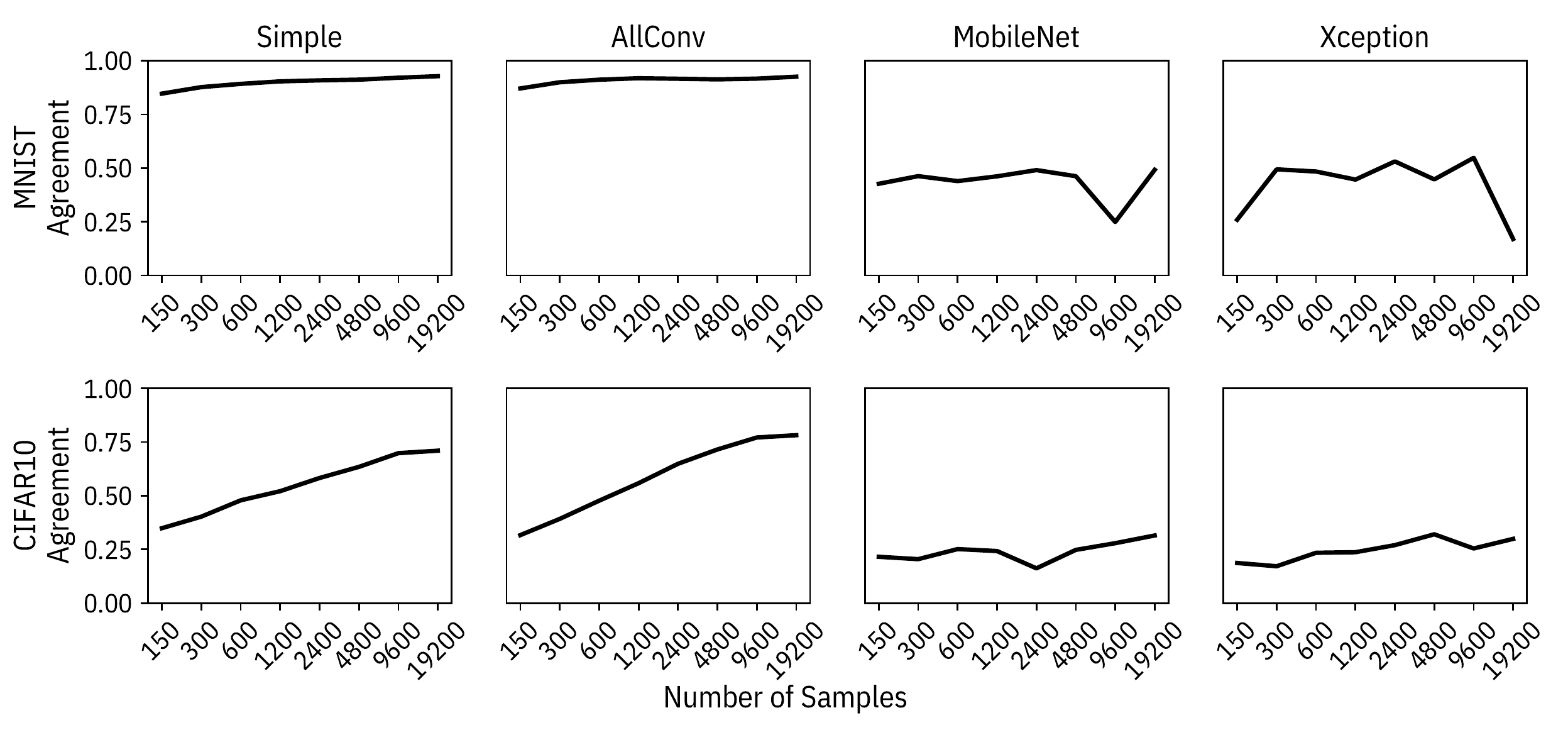}
  \caption{Agreement with the base model for various types of attacker model
  architectures.}
  \label{fig:attacker-model}
\end{figure*}


\paragraph{Model Architecture}
The attacker has a choice of their own models to train.
We test \sample{} to train
Simple model described above, as well as MobileNet~\cite{MobileNet},
AllConv~\cite{AllConvNet}, and Xception~\cite{Xception} models.
More complex models such ResNet~\cite{ResNet} and Inception~\cite{InceptionV3}
are not applicable to the test datasets we use due to the input image dimensions.

\fig~\ref{fig:attacker-model} shows comparisons of \underline{performance}
regarding different replicated model
architectures used.\footnote{Note that the performance degradation is
due to the use of only two third of the original training data less the attacker portion.}
As we can expect, the simplest model learns fastest.
While approaches such as Xception and MobileNet work well with larger data~\cite{MobileNet,Xception},
they are not suitable in a model stealing scenario with relatively few samples.
Also, we find AllConv not only learns relatively fast, but also performs well for all datasets.
Therefore, we use AllConv throughout the rest of the experiments. 
For the IMDB dataset, we use the most popular approach using Bidirectional LSTM~\cite{BiLSTM}.

\subsection{Defense Evaluation} \label{sec:exp:defense}


Now we compare defense methods against \sample{} attack method with AllConv model.
As a defense, we consider adding one of the following perturbations
to the output probability vector, where $\mathcal{U}(-1,1)$
is uniform random noise between -1 to 1, and $\beta$, $\rho$ and $f$ are parameters.
\begin{itemize}
\setlength\itemsep{0em}
\item \textbf{Uniform Random}: $\beta \mathcal{U}(-1,1)$.
\item \textbf{Uniform Random $\times$ Concave}: $\beta \mathcal{U}(-1,1) \times 1/e^{-\frac{1}{2} \left(y^i_j/\rho\right)^2}$.
\item \textbf{Uniform Random $\times$ Convex}: $\beta \mathcal{U}(-1,1) \times \left(1-1/e^{-\frac{1}{2} \left(y^i_j/\rho\right)^2} \right)$.
\item \textbf{Ranking-preserving Uniform Random}: Same as \textbf{Uniform Random}, but the ranking of output classes is preserved to maintain the accuracy.
\item \textbf{Sine}: $\beta \sin(f y^i_j)$.
\item \textbf{Reverse Sigmoid}: a stretched and reversed sigmoid $r(y_j^i)$ explained in this paper.
\end{itemize}

\begin{figure*}
  \centering
  \includegraphics[width=1\linewidth]{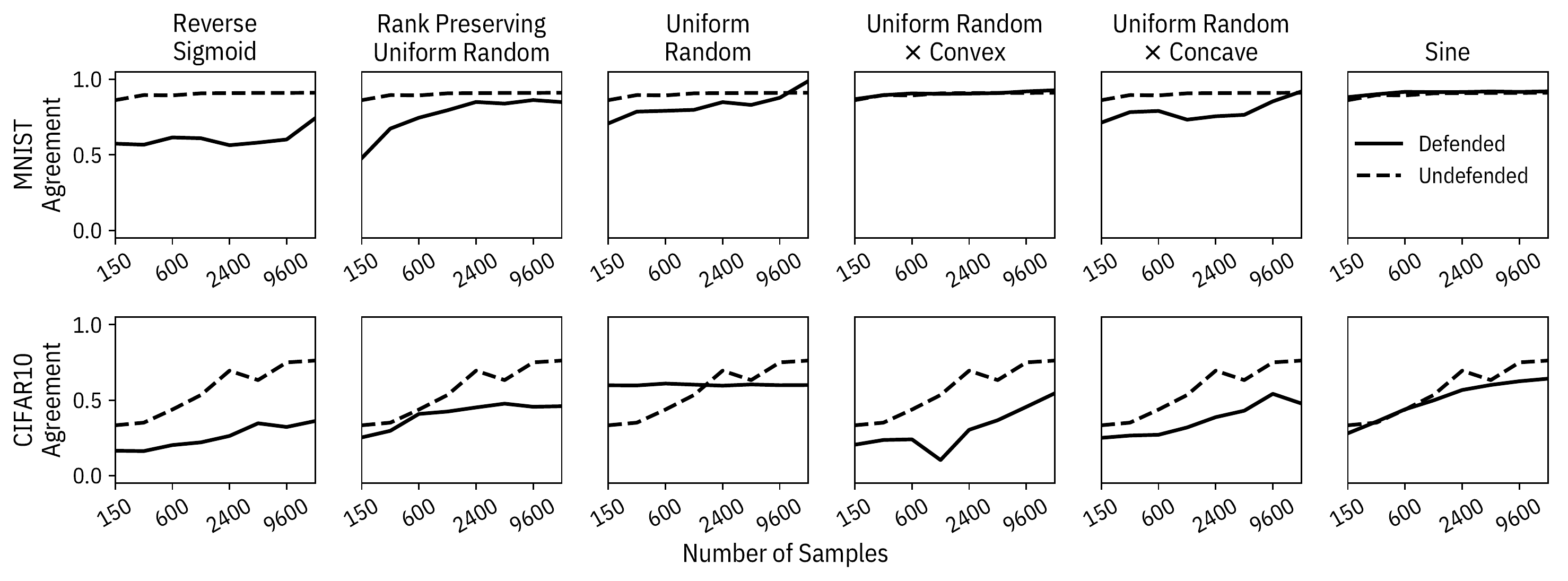}
  \caption{Agreement of stolen model with base model using different defense types.}
  \label{fig:defenses}
\end{figure*}

\begin{table*}[tb]
\centering
\caption{Agreements with 19200 queries. $M$: base model, $S(\cdot)$: stolen model, $P(\cdot)$: protected model.}\label{tab:exp:agreefive}
\scriptsize
\begin{tabular}{crlrlrlrlrlrl}
\toprule
Dataset & \multicolumn{2}{c}{MNIST} & \multicolumn{2}{c}{FASHION MNIST} & \multicolumn{2}{c}{CIFAR-10} & \multicolumn{2}{c}{CIFAR-100} & \multicolumn{2}{c}{STL-10} & \multicolumn{2}{c}{IMDB} \\ \hline
$S(M)$ & 0.93 & & 0.95 & & 0.82 & & 0.48 & & 0.62 & & 0.84 \\ \hline
$P(M, \text{Reverse Sigmoid)}$
	& 0.99 & \hspace{-4mm} \multirow{2}{*}{\rotatebox[origin=c]{-90}{$\curvearrowright$} \textbf{-0.31}} 
	& 0.99 & \hspace{-4mm} \multirow{2}{*}{\rotatebox[origin=c]{-90}{$\curvearrowright$} -0.29}
	& 0.85 & \hspace{-4mm} \multirow{2}{*}{\rotatebox[origin=c]{-90}{$\curvearrowright$} -0.45}
	& \textbf{1.00} & \hspace{-4mm} \multirow{2}{*}{\rotatebox[origin=c]{-90}{$\curvearrowright$} \textbf{-0.98}}
	& 0.99 & \hspace{-4mm} \multirow{2}{*}{\rotatebox[origin=c]{-90}{$\curvearrowright$} \textbf{-0.62}}
	& \textbf{1.00} & \hspace{-4mm} \multirow{2}{*}{\rotatebox[origin=c]{-90}{$\curvearrowright$} \textbf{-0.40}}
	\\
  $S(M, \text{Reverse Sigmoid})$ & \underline{0.68} & & 0.70 & & \underline{0.40} & & \underline{0.02} & & 0.37 & & 0.60 & \\ \hline
  $P(M, \text{Uniform Random}{})$
	& 0.60 & \hspace{-4mm} \multirow{2}{*}{\rotatebox[origin=c]{-90}{$\curvearrowright$} +0.38}
	& 0.56 & \hspace{-4mm} \multirow{2}{*}{\rotatebox[origin=c]{-90}{$\curvearrowright$} \textbf{-0.47}}
	& 0.79 & \hspace{-4mm} \multirow{2}{*}{\rotatebox[origin=c]{-90}{$\curvearrowright$} -0.13}
	& 0.82 & \hspace{-4mm} \multirow{2}{*}{\rotatebox[origin=c]{-90}{$\curvearrowright$} -0.80}
	& 0.55 & \hspace{-4mm} \multirow{2}{*}{\rotatebox[origin=c]{-90}{$\curvearrowright$} -0.43}
	& 0.72 & \hspace{-4mm} \multirow{2}{*}{\rotatebox[origin=c]{-90}{$\curvearrowright$} -0.12}
	\\
  $S(M, \text{Uniform Random}{})$ & 0.98 & & \underline{0.09} & & 0.66 & & \underline{0.02} & & \underline{0.12} & & 0.60 & \\ \hline
  $P(M, \text{Uniform Random} \times \text{Concave})$
	& 0.88 & \hspace{-4mm} \multirow{2}{*}{\rotatebox[origin=c]{-90}{$\curvearrowright$} +0.05}
	& 0.86 & \hspace{-4mm} \multirow{2}{*}{\rotatebox[origin=c]{-90}{$\curvearrowright$} +0.05}
	& 0.69 & \hspace{-4mm} \multirow{2}{*}{\rotatebox[origin=c]{-90}{$\curvearrowright$} -0.15}
	& 0.82 & \hspace{-4mm} \multirow{2}{*}{\rotatebox[origin=c]{-90}{$\curvearrowright$} -0.80}
	& 0.92 & \hspace{-4mm} \multirow{2}{*}{\rotatebox[origin=c]{-90}{$\curvearrowright$} -0.35}
	& 0.72 & \hspace{-4mm} \multirow{2}{*}{\rotatebox[origin=c]{-90}{$\curvearrowright$} -0.13}
	\\
  $S(M, \text{Uniform Random} \times \text{Concave})$ & 0.93 & & 0.91 & & 0.54 & & \underline{0.02} & & 0.57 & & \underline{0.59} & \\ \hline
  $P(M, \text{Uniform Random} \times \text{Convex})$
	& 0.89 & \hspace{-4mm} \multirow{2}{*}{\rotatebox[origin=c]{-90}{$\curvearrowright$} \textbf{+0.04}}
	& 0.86 & \hspace{-4mm} \multirow{2}{*}{\rotatebox[origin=c]{-90}{$\curvearrowright$} +0.04}
	& 0.70 & \hspace{-4mm} \multirow{2}{*}{\rotatebox[origin=c]{-90}{$\curvearrowright$} -0.12}
	& 0.82 & \hspace{-4mm} \multirow{2}{*}{\rotatebox[origin=c]{-90}{$\curvearrowright$} -0.80}
	& 0.98 & \hspace{-4mm} \multirow{2}{*}{\rotatebox[origin=c]{-90}{$\curvearrowright$} -0.39}
	& 0.72 & \hspace{-4mm} \multirow{2}{*}{\rotatebox[origin=c]{-90}{$\curvearrowright$} -0.12}
	\\
  $S(M, \text{Uniform Random} \times \text{Convex})$ & 0.93 & & 0.90 & & 0.58 & & \underline{0.02} & & 0.59 & & 0.60 & \\ \hline
  $P(M, \text{Ranking-preserving Uniform Random}{})$
	& \textbf{1.00} & \hspace{-4mm} \multirow{2}{*}{\rotatebox[origin=c]{-90}{$\curvearrowright$} -0.13}
	& \textbf{1.00} & \hspace{-4mm} \multirow{2}{*}{\rotatebox[origin=c]{-90}{$\curvearrowright$} -0.05}
	& \textbf{1.00} & \hspace{-4mm} \multirow{2}{*}{\rotatebox[origin=c]{-90}{$\curvearrowright$} \textbf{-0.54}}
	& \textbf{1.00} & \hspace{-4mm} \multirow{2}{*}{\rotatebox[origin=c]{-90}{$\curvearrowright$} -0.77}
	& \textbf{1.00} & \hspace{-4mm} \multirow{2}{*}{\rotatebox[origin=c]{-90}{$\curvearrowright$} -0.45}
	& \textbf{1.00} & \hspace{-4mm} \multirow{2}{*}{\rotatebox[origin=c]{-90}{$\curvearrowright$} -0.24}
	\\
  $S(M, \text{Ranking-pre. Uniform Random}{})$ & 0.87 & & 0.95 & & 0.46 & & 0.23 & & 0.55 & & 0.76 & \\ \hline
  $P(M, \text{Sine})$
	& 0.89 & \hspace{-4mm} \multirow{2}{*}{\rotatebox[origin=c]{-90}{$\curvearrowright$} +0.03}
	& 0.91 & \hspace{-4mm} \multirow{2}{*}{\rotatebox[origin=c]{-90}{$\curvearrowright$} -0.02}
	& 0.77 & \hspace{-4mm} \multirow{2}{*}{\rotatebox[origin=c]{-90}{$\curvearrowright$} -0.03}
	& 0.56 & \hspace{-4mm} \multirow{2}{*}{\rotatebox[origin=c]{-90}{$\curvearrowright$} -0.08}
	& 0.61 & \hspace{-4mm} \multirow{2}{*}{\rotatebox[origin=c]{-90}{$\curvearrowright$} +0.03}
	& 0.84 & \hspace{-4mm} \multirow{2}{*}{\rotatebox[origin=c]{-90}{$\curvearrowright$} -0.04}
	\\
  $S(M, \text{Sine})$ & 0.92 & & 0.89 & & 0.74 & & 0.48 & & 0.64 & & 0.80 & \\ \bottomrule
\end{tabular}
\end{table*}

The goal of the defense is twofold:
1) prevent model replication (low stolen model $S(M, \mathit{defense})$ agreement),
and 2) retain the performance of the protected model (high protected model $P(M, \mathit{defense})$ agreement).
The best parameters for the defense methods are searched using grid search
to achieve the highest cosine similarity to the original model
with at least 20\% accuracy drop in stealing with attacker budget of 19200.
When we cannot find parameters satisfying 20\% accuracy drop in stealing,
we instead choose the parameters with the highest accuracy drop in stealing.

\fig~\ref{fig:defenses} and \tab~\ref{tab:exp:agreefive} shows the experimental results.
We see the proposed Reverse Sigmoid defense
consistently slows down the replication process (\fig~\ref{fig:defenses}),
and can drop the stolen model agreement more than 20\% for all tested datasets (\tab~\ref{tab:exp:agreefive}).
Also, the protected models still have high agreements.
Using a sinusoidal wave noise does not protect the model
for all tested parameter combinations.
The final perturbation made by sine is small
as the sine applied to each class can interfere each other,
and it is normalized to have only small effect.
While \textbf{Uniform Random} may look effective
for some datasets like FASHION MNIST and CIFAR-100,
the agreements of the protected models are not reliable.
In case of \textbf{Ranking-preserving Uniform Random},
we can achieve the perfect agreements,
but the effectiveness of the defense is unpredictable (e.g., FASHION MNIST vs. CIFAR-10).

\begin{table*}[bt]
\centering
\caption{Effects of Reverse Sigmoid defense. $M$: base model, $P$: protected model, $S(P)$: stolen model from $P$.} \label{tab:defense:rs}
\small
\begin{tabular}[tb]{ccccccc}
\toprule
Dataset & $M$ acc. & $P$ acc. & $S(P)$ acc. & Cos($P$, $M$) & MAE($P$, $M$) & KL-div($P$, $M$) \\ \hline
MNIST & 0.99 & 0.99 & 0.68 & 0.41 & 0.17 & 3.27 \\ 
FASHION MNIST & 0.87 & 0.87 & 0.66 & 0.47 & 0.16 & 1.72 \\ 
CIFAR-10 & 0.75 & 0.73 & 0.36 & 0.61 & 0.48 & 1.26 \\ 
CIFAR-100 & 0.43 & 0.43 & 0.02 & 0.33 & 0.02 & 2.30 \\ 
STL-10 & 0.51 & 0.51 & 0.31 & 0.74 & 0.16 & 2.20 \\ 
IMDB & 0.81 & 0.81 & 0.60 & 0.80 & 0.39 & 0.48 \\
\bottomrule
\end{tabular}
\end{table*}

\tab~\ref{tab:defense:rs} shows the side effects of the Reverse Sigmoid defense.
The protected model accuracy is well maintained
because the defense tends not to change the labels if one class is dominating.
To measure how much noise is added to the probability vectors,
we use the average cosine similarity,
the mean absolute error per dimension, and KL-divergence
of the protected model and the base model outputs.
In particular, we see that the mean absolute error is less than 0.2 in most datasets.
This noise can be higher if the model prediction is more confident.
However, in this case, the decision is unlikely to be affected in many applications.

\subsection{Base Model Accuracy versus Reverse Sigmoid Protection Effectiveness} \label{sec:exp:add}

\begin{table}[hbt]
\centering
\caption{Base model and stealing on CIFAR-10} \label{tab:defense:arch}
\begin{tabular}[tb]{cccccc}
\toprule
Model & $M$ acc. & $P$ acc. & $S$ acc. & Acc. drop \\ \hline
AllConv & 0.75 & 0.73 & 0.36 & 0.37 \\
Simple & 0.70 & 0.69 & 0.50 & 0.19 \\
Xception & 0.36 & 0.36 & 0.23 & 0.13 \\
MobileNet & 0.24 & 0.22 & 0.21 & 0.01 \\
\bottomrule
\end{tabular}
\end{table}

Since we add noise to the confident classes, the output probability of a model
can affect the performance of the Reverse Sigmoid defense.
To see this, we train different models on CIFAR-10 dataset, resulting in different accuracies.
Then, we apply the Reverse Sigmoid protection, and try the attack with AllConv.
In \tab~\ref{tab:defense:arch}, we see that
if the original accuracy ($M$ accuracy) is higher,
the defense is more effective (higher accuracy drops).
This is especially important because
when the accuracy of the original model is high,
there is higher demand of protection.


\begin{figure*}[h]
  \centering
  \includegraphics[width=0.8\linewidth]{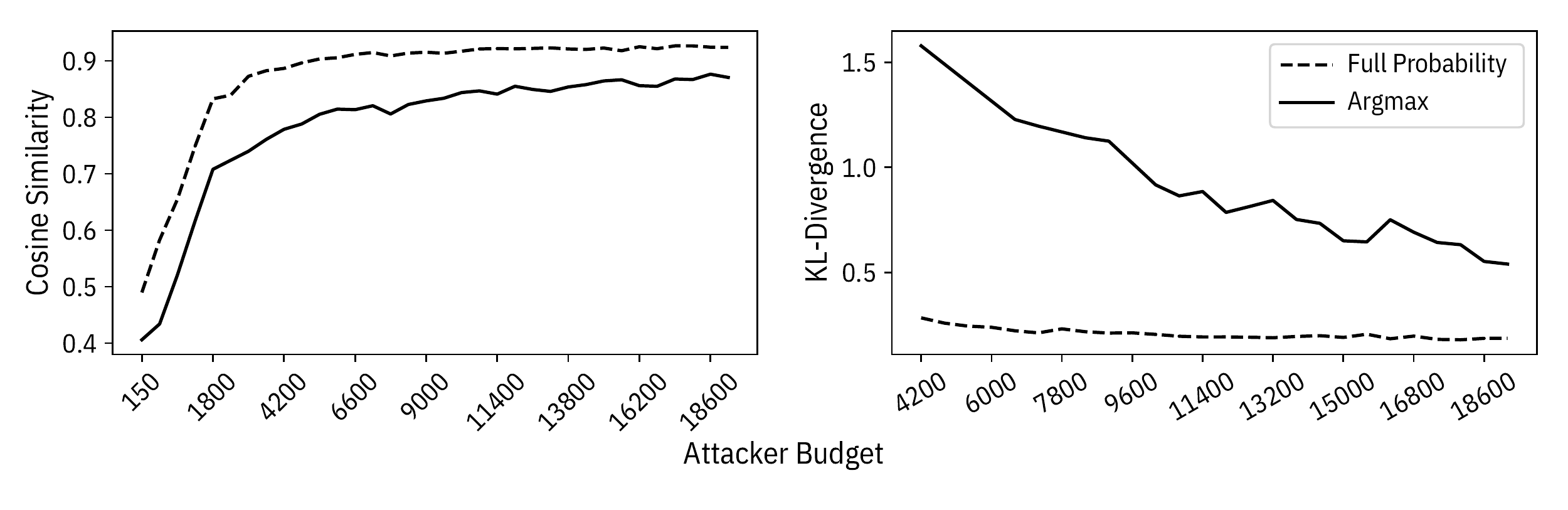}
  \caption{Performance difference of model stealing using
  all class probabilities vs.\ only the
  top-1 label on CIFAR-10.}
  \label{fig:argmax_compare}
\end{figure*}



\subsection{Robustness of Reverse Sigmoid against Defense-aware Attack} \label{sec:exp:variants}

We test the robustness of the Reverse Sigmoid defense when the attacker knows more information
on CIFAR-10 dataset.

\paragraph{Attack using the same defense layer}
If the defense and the parameters get to be known by the attacker,
the attacker can use exactly the same defense layer in their model.
The Reverse Sigmoid defense is not a standard neural network layer, and have ambiguity
that multiple logit values are mapped to the same probability by the defense layer.
This essentially propagates wrong gradient values to the model,
degrading the stealing process as resulted in
\tab~\ref{tab:defense:robustness}.

\paragraph{Attack using the mean-squared-error loss function}
To replicate the model output more precisely, the attacker may use
the mean-squared-error (MSE) loss function instead of the more common cross entropy loss function
which the Reverse Sigmoid defense is designed for.
However, for the same reason as the attack using the same defense layer,
this loss function is not free from the ambiguity,
and still shows poor performance as shown in \tab~\ref{tab:defense:robustness}.

\begin{figure}[t!]
	\centering\includegraphics[width=1.0\linewidth]{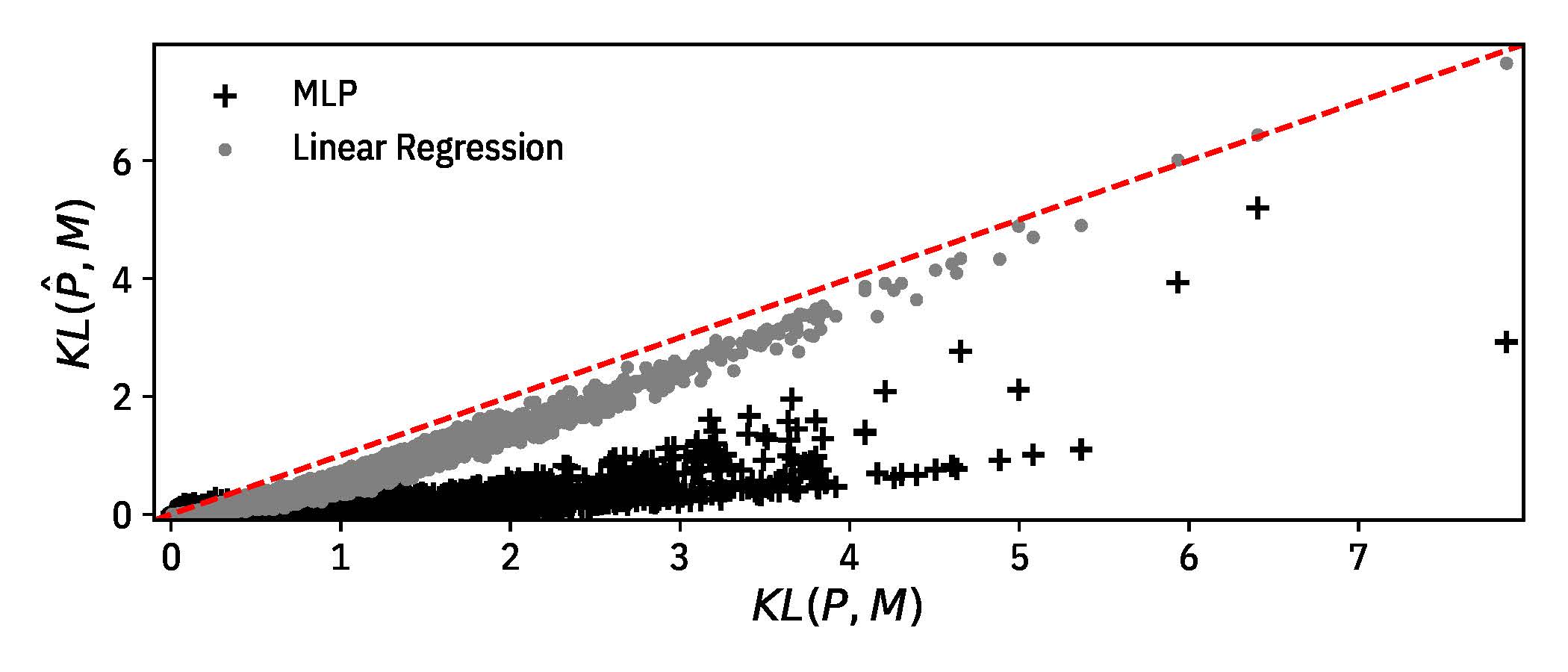}
	\caption{Recovering unprotected class probabilities using linear regression and MLP}\label{fig:attack:kl_diff}
\end{figure}

\paragraph{Attack using an inversion mapping}
We now evaluate the adversary's ability to recover the original unprotected class probabilities from the protected model
assuming they have \emph{full knowledge} of the parameters of the protection, \ie, $y^i,\hat y^i$ pairs.
We compute a linear regression of the class probabilities of the base model $M$ and
the protected model $P$,
and use the regression and normalization
to attempt to recover the unprotected probabilities.
In our second attack, we use a simple multilayer perceptron (MLP) model with two hidden layers ($5 \times K$ and $3 \times K$ neurons), and $K$ input and output values.
The model is trained on 16,670 \emph{real} output pairs from $M$ and $P$ ($y^i,\hat y^i$ pairs) and optimizing over the loss $\mathsf{KL}(\hat P, M)$.
We compute the KL divergence of both the recovered probability values ($\hat P$)
and the protected probability values ($P$) against
the unprotected model in \fig~\ref{fig:attack:kl_diff}, where a perfect attack corresponds to $\mathsf{KL}(\hat P, M) = 0$.
It is clear to see that both attacks are able to recover some of the information lost by the attack, but neither is able to fully recover the unprotected model output.
In the linear regression model, a fixed amount of information is recovered,
while the MLP is able to recover more information.
However the attack's capability diminishes with the amount of noise that was added to the sample.
Finally, we find the MLP attack does not increase the stolen model agreement,
as shown in \tab~\ref{tab:defense:robustness}.

\begin{table}
	\begin{center}
		\caption{Agreements of attack variants} \label{tab:defense:robustness}
	\begin{tabular}{cc}
		\toprule
		Attack & Agree. \\ \hline
		Same Defense Layer & 0.10 \\
		MSE Loss & 0.18 \\
		Inversion (MLP) & 0.22 \\
		Argmax & 0.78 \\ \bottomrule
	\end{tabular}
	\end{center}
	
	\label{label}
\end{table}

\paragraph{Attack using labels only (\argmaxsample{})}
As shown in \sec~\ref{sec:exp:attack},
using argmax requires a larger budget than attacks using the probabilities ($\approx 64 \times$ queries compared to \sample{}).
However, the attacker can simply take the top-1 class of the result
which is usually correct.
This approach achieves much better replication of a defended model
given enough budget as shown in \tab~\ref{tab:defense:robustness}.
Still, by forcing the attacker to discard probability values,
the attacker has to use $4\times$ as many queries in our evaluation on CIFAR-10 dataset.
That is, using probability values
requires 4800 queries to reach 0.7813 agreement,
but without probability values, 19200 queries have
to be used to reach the same agreement.
Besides the agreement,
we can see lower cosine similarity and high KL-divergence,
\fig~\ref{fig:argmax_compare},
even with 19200 queries, meaning that while top-1 decision is quickly learned,
the trait of the network including output distributions takes more queries to replicate.
This difference can limit the generalization power,
and it can be especially important if the attacker further wants
to use the model for adversarial example generation~\cite{papernot2017jacobian}.
Finally, note that this is a degenerate attack that applies to any defense that maintains top-1 accuracy.

\section{Conclusion} \label{sec:con}

Neural networks are becoming one of the key assets of an enterprise,
but they are vulnerable to stealing attacks.
We proposed a method that can be applied to wide variety of neural network models
and evaluated the protection performance
over five datasets, four neural network architectures,
and diverse threat models and attack parameters.
Our approach either prevented the stealing entirely
or slowed down the stealing process up to $64\times$
in the worst case when the attacker knows the defense.

\bibliographystyle{aaai}
\bibliography{biblio}

\end{document}